\documentclass[10pt,twocolumn,letterpaper]{article}

\usepackage{cvpr}              
\usepackage{setspace}
\usepackage{multirow}
\usepackage{bm}
\usepackage[dvipsnames]{xcolor}
\newcommand{\red}[1]{{\color{red}#1}}

\definecolor{cvprblue}{rgb}{0.21,0.49,0.74}
\usepackage[pagebackref,breaklinks,colorlinks,citecolor=cvprblue]{hyperref}
\newcommand\blfootnote[1]{%
\begingroup
\renewcommand\thefootnote{}\footnote{#1}%
\addtocounter{footnote}{-1}%
\endgroup
}

\def\paperID{7890} 
\def\confName{CVPR}
\def\confYear{2024}

\title{FreePoint: Unsupervised Point Cloud Instance Segmentation}

\author{Zhikai Zhang$^{1}$,
Jian Ding$^{1,2}$$^{\dagger}$, Li Jiang$^{3}$, Dengxin Dai$^{4}$, Guisong Xia$^{1}$$^{\dagger}$\\
$^{1}$Wuhan University \quad $^{2}$KAUST \quad $^{3}$CUHK-Shenzhen \quad $^{4}$Huawei Zurich Research Center
}

\begin{document}
\maketitle
\blfootnote{$^\dagger$Corresponding author}

\begin{abstract}
Instance segmentation of point clouds is a crucial task in 3D field with numerous applications that involve localizing and segmenting objects in a scene. However, achieving satisfactory results requires a large number of manual annotations, which is a time-consuming and expensive process. To alleviate dependency on annotations, we propose a novel framework, FreePoint, for underexplored unsupervised class-agnostic instance segmentation on point clouds. In detail, we represent the point features by combining coordinates, colors, and self-supervised deep features. 
Based on the point features, we perform a bottom-up multicut algorithm to segment point clouds into coarse instance masks as pseudo labels, which are used to train a point cloud instance segmentation model. We propose an id-as-feature strategy at this stage to alleviate the randomness of the multicut algorithm and improve the pseudo labels' quality.
During training, we propose a weakly-supervised two-step training strategy and corresponding losses to overcome the inaccuracy of coarse masks. 
FreePoint has achieved breakthroughs in unsupervised class-agnostic instance segmentation on point clouds and outperformed previous traditional methods by over 18.2\% and a competitive concurrent work UnScene3D by 5.5\% in AP. Additionally, when used as a pretext task and fine-tuned on S3DIS, FreePoint performs significantly better than existing self-supervised pre-training methods with limited annotations and surpasses CSC by 6.0\% in AP with 10\% annotation masks. \textit{Code will be released at \url{https://github.com/zzk273/FreePoint}.}

\end{abstract}    
\newcommand{\jian}[1]{\textcolor[rgb]{0.08, 0.38, 0.74}{\textbf{Jian:} #1}}

\section{Introduction}
\label{sec:intro}
Instance segmentation on point clouds aims to segment and recognize objects in a 3D scene, serving as the foundation for a wide range of applications such as autonomous driving, virtual reality, and robot navigation. This task has received increasing attention~\cite{chen2021hierarchical,elich20193d,engelmann20203d,han2020occuseg,he2021dyco3d,jiang2020pointgroup,lahoud20193d,liang2021instance,yang2019learning,schult2022mask3d,vu2022softgroup} for the availability of large-scale point cloud datasets~\cite{armeni20163d,dai2017scannet,mo2019partnet,song2017semantic}. Most of the previous works focus on fully-supervised point cloud segmentation, which requires a large number of bounding boxes and per-point annotations to achieve satisfactory results. However, the annotations of point clouds are labor-intensive. For example, labeling an average scene in ScanNet takes about 22.3 minutes~\cite{dai2017scannet}. 

\begin{figure}[!t]
\centering
\vspace{10mm}
\includegraphics[width=1\linewidth]
{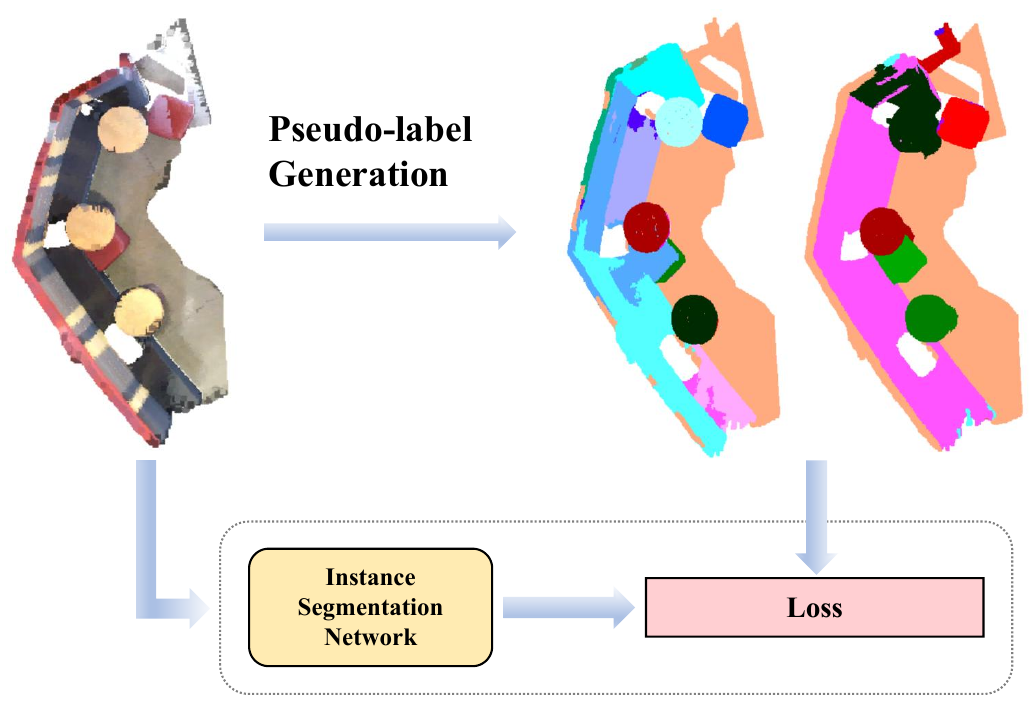}
\vspace{-0.5em}
\caption{We propose a novel framework for unsupervised point cloud instance segmentation. In detail, we cluster points based on coordinates, colors, and self-supervised deep features. Then we use the clustered pseudo masks to perform a step-training and improve the unsupervised segmentation quality further. }

\label{fig:demo}
\vspace{-1.0em}
\end{figure}

To relieve the annotation requirements, some weakly-supervised 3D segmentation methods~\cite{Xu20CVPR, Liu21CVPR, Zhang21AAAI, Zhang21ICCV,chibane2022box2mask} and semi-supervised 3D segmentation methods~\cite{Cheng21AAAI,Jiang2021CVPR} have been proposed. Besides, some works explore unsupervised pre-training methods for 3D point clouds~\cite{Hou21CVPR, xie2020pointcontrast,zhang2021self}, mainly focusing on data-efficient scene understanding and achieving satisfactory results when fine-tuning on downstream tasks with limited annotations. 
These works, however, still rely on considerable box, point annotations, or a certain proportion of mask annotations to achieve competitive results. A concurrent work Unscene3D~\cite{rozenberszki2023unscene3d} explores unsupervised 3D class-agnostic instance segmentation for indoor scenes. It shows promising results while still having large room for improvement in accuracy.

In this work, we propose a novel framework FreePoint for unsupervised point cloud instance segmentation, which can be split into three parts: (1) preprocessing and point feature extraction; (2) pseudo mask label generation by point feature based graph partitioning; (3) step-training using the pseudo labels. 
We first adopt plane segmentation algorithm repeatedly to split a point cloud scene into foreground points and background points. Then, for foreground points, we use a self-supervised pre-trained backbone to generate deep-learning feature embeddings for each point. To enhance our feature representation, we add coordinates and colors as extra point features. Our main motivation is that the geometry and color features are helpful for point cloud segmentation. These information has been widely adopted by some traditional point-clustering methods~\cite{rusu2010semantic, rusu20113d, achanta2012slic, papon2013voxel}.
To generate pseudo mask labels, we solve a bottom-up multicut~\cite{chopra1993partition} problem based on the affinities of point features and constructed point graphs. We propose an \textit{id-as-feature strategy} at this stage to alleviate the randomness of the multicut algorithm and improve the pseudo labels' quality. This strategy is, in essence, an ensemble of multiple runnings of RAMA. We also adopt down-sampling and up-sampling here to make the computation affordable. These pseudo masks are used to train an existing instance segmentation model.
In our work, we choose Mask3D~\cite{schult2022mask3d} for its efficiency and good performance.
Since the pseudo masks are inaccurate and the training can be unstable, we propose a weakly-supervised two-step training strategy and corresponding losses to alleviate this problem. The overview of FreePoint is shown in Figure~\ref{fig:demo}. 

We evaluate our method on \textit{unsupervised class-agnostic instance segmentation}. In this setting, our method shows surprising results without any annotations, surpassing previous SOTA by a large margin.
Apart from directly acquiring the class-agnostic instance masks, our method can also be used for unsupervised pre-training on 3D point clouds. The learned parameters of the backbone can be used to initialize a supervised instance segmentation model and improve final results with limited annotations.

Our contributions in this paper are three-fold:

\begin{itemize}
\item We propose a novel framework, FreePoint, for unsupervised point cloud instance segmentation with deep networks. Freepoint generates pseudo labels based on solving a graph partitioning problem and then uses these pseudo labels to train a 3D instance segmentation model. Our work opens up possibilities for advancing the field.

\item We make great efforts to overcome many difficulties brought by the lack of manual annotations. To generate pseudo labels of higher quality, we first propose a \textit{hybrid feature representation} for point affinity computation. Then we design an \textit{id-as-feature strategy} to alleviate the randomness of the graph partitioning method. For better use of the noisy pseudo labels, we further propose a carefully designed \textit{two-step training strategy and corresponding losses} to overcome pseudo labels' noise.

\item We evaluate FreePoint's performance on unsupervised class-agnostic point cloud instance segmentation. It surpasses traditional unsupervised segmentation methods by over 18.2\%, and even outperforms the competitive concurrent work UnScene3D~\cite{rozenberszki2023unscene3d} by 5.5\% in AP. We also evaluate FreePoint's performance as a pretext task. For example, when fine-tuning on S3DIS dataset with 10\% labeled masks, FreePoint outperforms training from scratch by +8.2\% AP and CSC by 5.8\% AP.

\end{itemize}

\section{Related work}
\label{sec:related_work}

\paragraph{Point cloud instance segmentation} 
Early works on point cloud instance segmentation focus on grouping points based on their affinities~\cite{Wang18CVPR, Wang19CVPR,elich20193d}. They use dense labels to train point feature encoders and segment point clouds by measuring the point affinities. 3D-SIS~\cite{hou20193d} and 3D-BoNet~\cite{yang2019learning} extract bounding box proposals and classify them. Recent works prefer to group points based on predicted semantics and object centers~\cite{jiang2020pointgroup, engelmann20203d, chen2021hierarchical, liang2021instance, han2020occuseg}. Mask3D~\cite{schult2022mask3d} is the first Transformer-based approach to challenge this task. We choose it as our step-training model for its high efficiency. The above works highly rely on per-point labels to achieve good results. However, acquiring such labels is labor-intensive.
Some 3D instance segmentation works have been proposed these years to alleviate dependency on costly manual annotations.~\cite{Xu20CVPR, Hou21CVPR, Cheng21AAAI, Jiang2021CVPR, Liu21CVPR, Xu20CVPR, Zhang21AAAI, Zhang21ICCV} assume a sparse number of points is annotated and~\cite{chibane2022box2mask} use only bounding box labels. However, they still rely on considerable annotations to achieve competitive results. 

\paragraph{Unsupervised segmentation and detection}
In 2D images, several works explore unsupervised object detection~\cite{cho2015unsupervised,simeoni2021localizing,wang2022tokencut,melas2022deep,shin2022unsupervised}, instance segmentation~\cite{wang2022freesolo,wang2023cut}, and semantic segmentation~\cite{Ke_2022_CVPR, cho2021picie,van2021unsupervised}. In object detection area, some works~\cite{wang2022tokencut, melas2022deep,shin2022unsupervised} use spectral methods to discover and segment main objects in a scene. They first construct an adjacency matrix using spatial features, color features, or features from pre-trained backbones. Then the matrix's eigenvectors and eigenvalues are computed to decompose the image. Recently, a few works~\cite{wang2022freesolo,wang2023cut} have explored unsupervised instance segmentation for 2D images and achieved satisfactory results. UnScene3D~\cite{rozenberszki2023unscene3d} has explored unsupervised 3D instance segmentation for indoor scenes. It operates on a basis of geometric oversegmentation to generate pseudo labels and refines them through self-training as many 2D works. UnScene3D shows promising results while still having large room for improvement in accuracy. The main difference between our method and this work lies in: (1) utilizing only 3D color and geometric features instead of multimodal features from 2D and 3D pre-training backbones; (2) designing a two-step training strategy instead of a multi-round self-training strategy which is very time-costing.

\paragraph{3D feature representation}
Traditional methods~\cite{achanta2012slic, papon2013voxel} use features like \textit{coordinates}, \textit{colors} and \textit{normals} to describe each point in a scene. Following the tendency of unsupervised pre-training in 2D field, various works~\cite{afham2022crosspoint,xie2020pointcontrast,Hou21CVPR,pang2022masked,yu2022point,
min2022voxel, zhang2021self,zhang2022point} have been proposed recently to represent 3D features, but mostly focusing on single-object classification tasks on ShapeNet~\cite{chang2015shapenet} or ModelNet~\cite{wu20153d}. Only a few works~\cite{xie2020pointcontrast, Hou21CVPR, zhang2021self} focus on large-scale indoor point cloud datasets, which are important for multi-object segmentation tasks and contain far more than only one object.~\cite{Hou21CVPR} mainly explores how to address downstream tasks in a data-efficient semi-supervised way rather than using full annotations. As a result, many works on instance segmentation and semantic segmentation train their model from scratch and can not benefit from 3D pre-training.
\section{Method}
\begin{figure*}[t!]
\centering
\includegraphics[width=0.90859095\linewidth]
{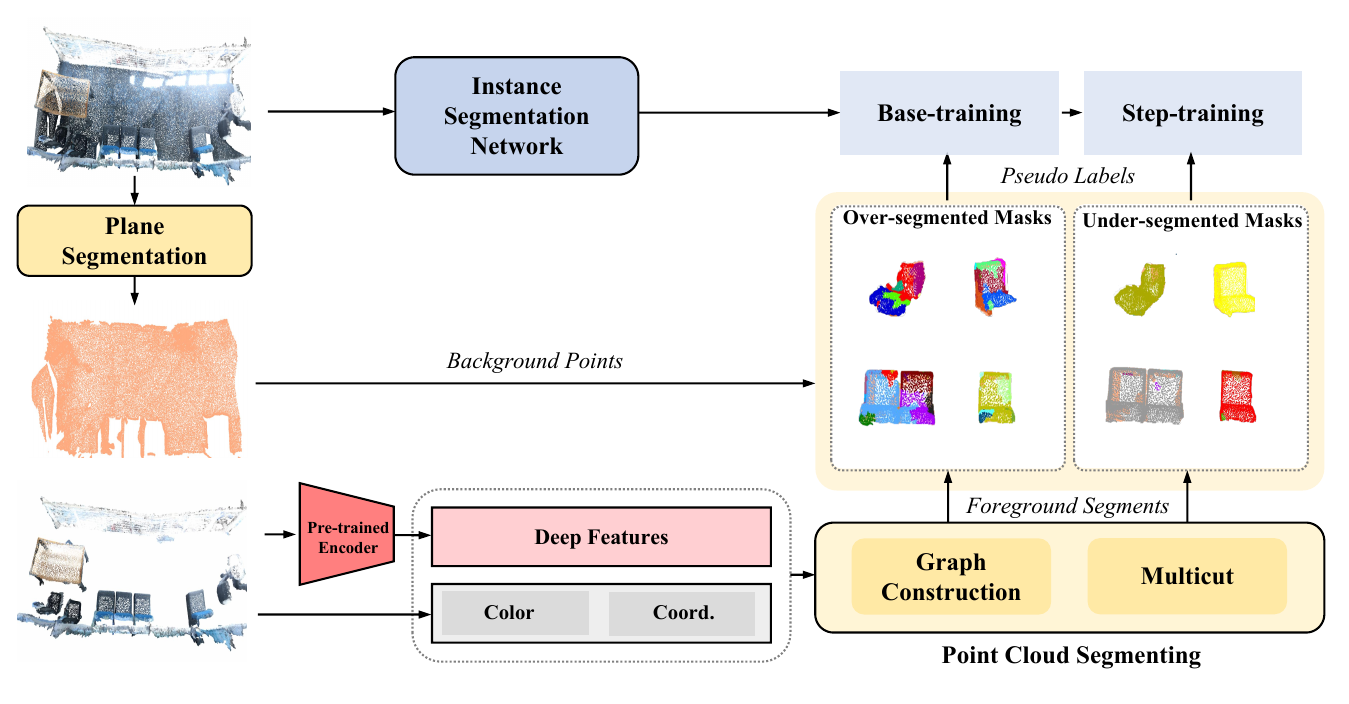}
\vspace{-0.5em}
\caption{\textbf{Overview.} For inputted point clouds, we first use plane segmentation to filter out backgrounds. Then we represent the features for points by combining self-supervised deep features and traditional features. After that, we construct a graph and compute the edge affinity costs between points. Based on the graph, we apply a multicut algorithm to segment point clouds into coarse instance masks. These masks are adopted as pseudo labels to train a 3D instance segmentation model with our proposed weakly-supervised loss and step-training strategy.
}
\label{fig:pipeline}
\vspace{-1.0em}
\end{figure*}
Our pipeline, as shown in Figure~\ref{fig:pipeline}, can be split into three parts: (1) preprocessing and point feature extraction; (2) pseudo mask label generation by point feature based graph partitioning; (3) step-training using the pseudo labels. 
Concretely, we first apply plane segmentation to separate the foreground points and background points. 
Then, for foreground points, we combine both traditional features (\textit{i.e.,} coordinates and colors) and self-supervised deep-learning embeddings to represent their features. Based on it, we construct an undirected graph ${\bf{G}} = ({\bf{V}}, {\bf{E}}, {\bf{A}})$ viewing the points as vertices ${\bf{V}}$ and their connections as edges ${\bf{E}}$. ${\bf{A}}$ is an affinity cost vector measured by the affinities between point features. After this, a multicut algorithm is adopted to decompose ${\bf{G}}$ into coarse instance masks. Finally, we use the coarse masks to perform step-training with our proposed weakly-supervised loss and step-training strategy. 

\subsection{Preprocessing and point feature extraction}
\label{feature_extraction}

\paragraph{Preprocessing}

It is difficult to directly cluster the point clouds into instance masks and backgrounds in the unsupervised setting, since numerous inconspicuous objects are integrated into nearby backgrounds.
However, we find that for indoor point cloud datasets, backgrounds include floors, walls, and ceilings, which are usually large and flat surfaces and thus can be easily removed. So we apply plane segmentation~\cite{zhou2018open3d} to filter out major surfaces in a scene and consider them as backgrounds. In detail, we run a non-deep learning plane segmentation algorithm several times for a scene. Each fitted plane will be projected and compared with its corresponding surface of the whole indoor scene's bounding box and we will compute the IOU. If the IOU is larger than a threshold, it will be seen as part of the background and removed from the scene. After this step, the original input point cloud 
${{\bf{V}}_{full}}\in{\mathbb{R}^{N \times6}}$, which contains coordinate and color information, is divided into two subsets: foreground point cloud ${{\bf{V}}_{fg}}\in{\mathbb{R}^{N_{fg} \times6}}$ and background point cloud ${{\bf{V}}_{bg}}\in{\mathbb{R}^{N_{bg} \times6}}$. Since segmenting backgrounds is not the goal of instance segmentation, we only use ${{\bf{V}}_{fg}}$ for the next feature extracting and point cloud segmenting step.

We then perform farthest point sampling~\cite{qi2017pointnet++} to down sample ${{\bf{V}}_{fg}}$ into ${{\bf{V}}_{sampled}}\in{\mathbb{R}^{N_{sampled} \times6}}$. This step is important for that: (1) it can reduce the computation cost of the following point cloud segmenting process and make the pseudo label generation on raw points affordable; (2) this down-sampling can make the point distribution more sparse. Because of the sparsity, the sampled points are farther from each other in feature space, which is beneficial for our point cloud segmenting method described in Section~\ref{segmenting}. 

\paragraph{Feature extraction} Since our segmenting method is based on the affinity between the feature representation of each point, we should find a way to make points closer in feature embedding space if they belong to the same object and farther otherwise. We first use self-supervised pre-trained backbones to encode points. 
However, we find it difficult to encode points discriminatively using deep-learning features alone, which means even points belonging to different instances can be close to each other in the feature embedding space. 

Before the era of deep learning, some methods~\cite{achanta2012slic, papon2013voxel} use traditional features to cluster points. For example, Supervoxel~\cite{papon2013voxel} uses features like \textit{coordinates} and \textit{colors} to measure the affinities between points and cluster them accordingly. Inspired by it, we use both traditional features and deep-learning features to represent each sampled point and measure their affinities in our work.

\subsection{Point cloud segmenting}
\label{segmenting}

\paragraph{Preliminary} Minimum-cost multicut~\cite{chopra1993partition} problem aims to decompose an undirected graph ${\bf{G}} = ({\bf{V}}, {\bf{E}}, {\bf{A}})$ into a set of point subsets ${\{{\bf{V}}_{1},\ldots,{\bf{V}}_{k}\}}$ where ${\bf{V}}_1 \cup \ldots \cup {\bf{V}}_k = {\bf{V}}$ and ${\bf{V}}_i \cap {\bf{V}}_j = \varnothing$ $\forall i\neq j$. Edges that straddle distinct clusters which decomposes ${\bf{G}}$ form the \emph{cut} $\delta({\bf{V}}_1,\ldots,{\bf{V}}_k)$. ${\bf{A}} \in \mathbb{R}^{\bf{E}}$ is an affinity cost vector. Each edge $(u,v) \in {\bf{E}}$ has a cost ${\bf{A}}_{(u,v)}$. We need to find a decomposition \emph{cut} of the undirected graph ${\bf{G}}$ that agrees as much as possible with the affinity cost vector, minimizing the whole cost of \emph{cut}. So if more edge cost values are negative, ${\bf{G}}$ will be decomposed into more clusters generally.  

\paragraph{Segmenting}In our work, we select RAMA~\cite{abbas2022rama}, a rapid bottom-up multicut algorithm on GPU, to segment point clouds. Each $v_{i} \in {\bf{V}}$ is connected to the closest $k_{1}$ points ${\{u_{i_{1}}, . . . , u_{i_{k_{1}}}\}} \in {\bf{V}}$ by edges $(v_i,u_{i_{j}}) \in {\bf{E}}$, where $j \in \{1,\ldots,k_{1}\}$.
Affinity cost vector ${\bf{A}}$ is the affinities of both deep features and traditional features. 
For deep-learning feature embeddings ${{\bf{F}}}\in{\mathbb{R}^{N_{sampled} \times {dim}}}$, we calculate their cosine similarities: 
\begin{equation}
\begin{aligned}
\mathcal{\bf A}_{(i,j),emb} = {\tt Cos}({\bf F}_{i}, {\bf F}_{j}).
\end{aligned}
\end{equation}

We split ${{\bf{V}}_{sampled}}$ into point coordinates ${{\bf{P}}}\in{\mathbb{R}^{N_{sampled} \times3}}$ and point colors ${{\bf{C}}}\in{\mathbb{R}^{N_{sampled} \times3}}$. Then we compute L2 distance respectively in XYZ space and RGB space:
\begin{equation}
\begin{aligned}
\mathcal{\bf A}_{(i,j),xyz} = -\parallel{\bf P}_{i}, {\bf P}_{j}\parallel_2, \mathcal{\bf A}_{(i,j),rgb} = -\parallel{\bf C}_{i}, {\bf C}_{j}\parallel_2.
\end{aligned}
\end{equation}

These three affinities are all normalized to have a mean value of 0 and variance of 1. The total affinity can be written as:
\begin{equation}
\label{eq:loss_mask}
\begin{aligned}
\mathcal{\bf A} &= \alpha_{1}{\bf A}_{emb} + \alpha_{2}{\bf A}_{xyz} + \alpha_{3}{\bf A}_{rgb},
\end{aligned}
\end{equation}
where $\alpha_{1}$, $\alpha_{2}$, $\alpha_{3}$ are the weights to balance the importance of different affinities. ${\bf{G}}$ will be sent to RAMA~\cite{abbas2022rama} based on ${\bf{A}}$ and the output is pseudo instance labels.

However, due to the characteristics of this bottom-up segmenting method, the generated coarse masks have randomness. We design an id-as-feature strategy to solve this problem and improve the pseudo labels' quality. This strategy is, in essence, an ensemble of multiple generation results. Concretely, each time $t$ we run RAMA, every point in ${{\bf{V}}_{sampled}}$ will have an assigned pseudo instance label $id_{t}$. We run RAMA multiple times ${\bf{T}}$ and concatenate every $id_{t}$ to form a new feature for each point in ${{\bf{V}}_{sampled}}$. For these id-generated features ${\bf{IDF}}\in{\mathbb{R}^{N_{sampled} \times {T}}}$, their similarities will be computed as:
\begin{equation}
\label{eq:loss_mask}
\begin{aligned}
\mathcal{\bf{A}}_{(i,j),id} &= \frac{1}{T}\sum_{t = 1}^T\boldsymbol{I}\left[ {\bf{IDF}}_{i}\left[t\right] = {\bf{IDF}}_{j}\left[t\right]\right]
\end{aligned}
\end{equation}

\begin{figure}[t]
\begin{center}
\includegraphics[width=1.0\linewidth]{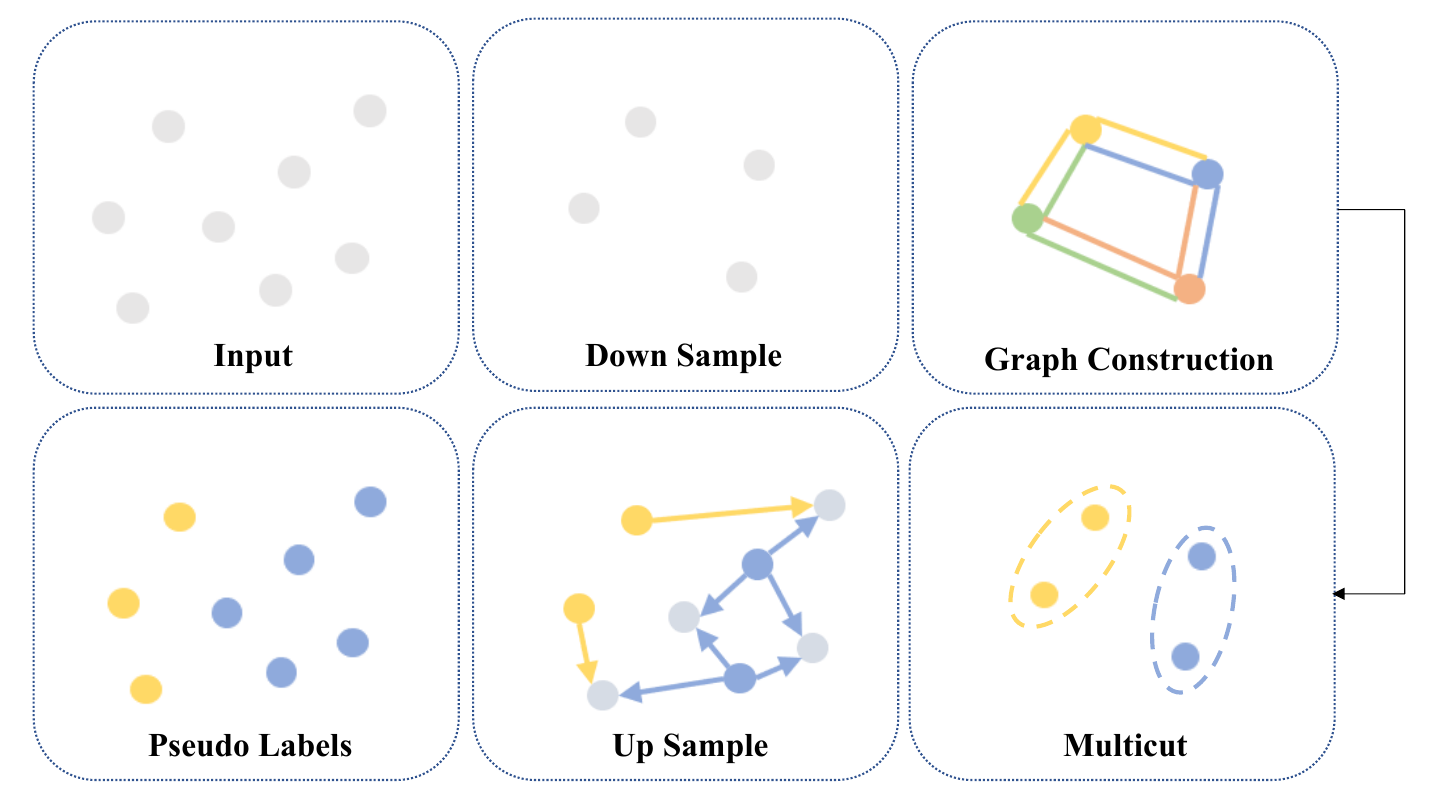}
\end{center}
   \caption{\textbf{Pseudo-label Generation.} In this figure, we show the complete pipeline of pseudo-label generation. For simplicity, we set $k_{1} = k_{2} = 2$. }
\label{fig:pseudo_label generation}
\end{figure}

We run RAMA again based on ${\bf A}_{id}$ and then preliminary pseudo instance labels ${{\bf{L}}_{sampled}}\in{\mathbb{R}^{N_{sampled}}}$ are formed. To recover to original size, we use $knn$ to find the closest $k_{2}$ points and corresponding labels in ${{\bf{V}}_{sampled}}$ for each point in ${{\bf{V}}_{fg}}$. By majority voting of the $k_{2}$ points, we obtain ${{\bf{L}}_{fg}}\in{\mathbb{R}^{N_{fg}}}$. 
Then we annotate points in ${{\bf{V}}_{bg}}$ as background and concatenate it with ${{\bf{L}}_{fg}}$ to obtain final pseudo labels ${{\bf{L}}}\in{\mathbb{R}^{N}}$. The pipeline of pseudo-label generation is shown in Figure~\ref{fig:pseudo_label generation}.

As mentioned before, RAMA generally segments the scene into more objects if more edge values are negative. When running RAMA based on ${\bf{A}}$, we will add different hyper-parameters $\sigma_{low}, \sigma_{high}$ to affinity: 
\begin{equation}
\label{eq:loss_mask}
\begin{aligned}
\mathcal{\bf A}_{final} &= {\bf A} + \{\sigma_{low}, \sigma_{high}\}. 
\end{aligned}
\end{equation}

By changing $\sigma$, we generate coarse masks of two different segmenting levels. One is able to localize and identify most objects in the scene but fails to generate complete masks for instances. We denote these masks as base masks. To overcome base masks' defects, we generate under-segmented masks with a relatively higher $\sigma$. They will be in good use for the next step following our weakly-supervised two-step training design. It is worth mentioning that when running RAMA based on ${\bf{A}}_{id}$, $\sigma$ is automatically chosen to keep the number of generated instances approximately the same as the average instance number of ${\bf{T}}$ runnings of RAMA based on ${{\bf{A}}_{final}}$.

\subsection{Training with coarse masks}
\label{coarsemask}

To further refine the coarse masks, we aim to train a point cloud instance segmenter using these masks as pseudo labels. In our work, we choose Mask3D~\cite{schult2022mask3d}, a Transformer-based model for semantic instance segmentation, for its good performance and efficiency. Coarse masks are often inaccurate, so directly using them to train an instance segmenter in a fully-supervised way will cause unsatisfactory results. Therefore we propose two designs to solve this problem, including a new weakly-supervised loss and a step-training strategy.

\paragraph{Loss for weakly-supervised training}
In the original implementation of Mask3D~\cite{schult2022mask3d}, they use both dice loss $\mathcal{L}_{dice}$ and binary cross entropy loss $\mathcal{L}_{BCE}$ as mask loss to train. However, our pseudo labels are inaccurate, so using such per-point loss directly may lead to sub-optimal results. We propose to use these coarse masks as a kind of weak annotation and design a weakly-supervised loss.

Inspired by ~\cite{wang2022freesolo,chibane2022box2mask,tian2021boxinst}, we believe mask centers and bounding boxes are important for weakly-supervised training. Mask centers can help to localize instances. We compute the mean value of normalized coordinates in a predicted mask ${\bf m}$ and target mask ${\bf m^{*}}$ along each axis to get prediction center ${c_{mean}}\in(x_{c}, y_{c}, z_{c})$ and target center ${t_{mean}}\in(x_{t}, y_{t}, z_{t})$. Our model is trained to minimize the Euclidean distance between $c_{mean}$ and $t_{mean}$:
\begin{equation}
\begin{aligned}
\mathcal{L}_{mean} =   {\tt Euclidean}({avg}({\tt {\bf m}}), {avg}({\tt {\bf m^{*}}})). 
\end{aligned}
\end{equation}

We further propose a bounding box loss. Bounding box supervision enforces predictions with the correct sizes and locations. This design can further improve our work's performance. For implementation, we pick the maximum and minimum value along each axis for a predicted mask and a target mask to get two boundary point pairs $(c_{max}, t_{max})$ and $(c_{min}, t_{min})$. 
The Euclidean distance of each pair is summed to be our bounding-box loss. The loss can be written as:
\begin{equation}
\begin{aligned}
\mathcal{L}_{box} =   {sum}(&{\tt Euclidean}({max}({\tt {\bf m}}),
{{max}(\tt {\bf m^{*}})}),
\\
&{\tt Euclidean}({min}({\tt {\bf m}}), {{min}(\tt {\bf m^{*}})})).
\end{aligned}
\end{equation}

We compute the above losses directly on points without voxelization. Then the weighted sum of each term in weakly-supervised loss and fully-supervised loss will be our final loss, which can be written as:
\begin{equation}
\label{eq:loss_mask}
\begin{aligned}
\mathcal{L} &=  \lambda_{dice}\mathcal{L}_{dice} + \lambda_{BCE}\mathcal{L}_{BCE} 
\\
&+ \lambda_{mean}\mathcal{L}_{mean} + \lambda_{box}\mathcal{L}_{box},
\end{aligned}
\end{equation}
where $\lambda_{dice}$, $\lambda_{BCE}$, $\lambda_{mean}$, and $\lambda_{box}$ are the weights to balance the importance of different loss terms.

\paragraph{Step training strategy}
In section~\ref{segmenting}, we observe that our segmenting method can generate masks of different segmenting levels. For base masks, the scene will be generally split into object parts. More instances can be identified and localized in this situation, but they lack complete masks. For the under-segmented setting, the scene has fewer instance proposals, which means we will have more masks covering a whole object. However, instances in this setting are always mistakenly connected with nearby instances especially when they share similar features.

We wonder which kind of masks we should use to achieve better results. Both coarse masks have insurmountable defects if adopted as pseudo labels alone. Therefore, we explore a novel training strategy so that over-segmented and under-segmented masks can compensate for each other's shortcomings and significantly improve final results. Concretely, we use base masks as pseudo labels for the first training step. At this stage, the model is trained to segment points of similar features, 
regardless of whether they belong to object parts or whole objects.
For the second training step, we use under-segmented masks instead. With only a few epochs, the model learns to connect mistakenly segmented object parts into a whole object. This step can improve the results of the first step by a large margin with little time cost. 

However, the under-segmented masks which contain multiple objects may harm the model's performance. At this stage, we propose an undersegmentation-ignore design to relieve this problem. Concretely, during the bipartite matching stage of the training of Mask3D, we ignore the match if the matched pseudo mask contains more than certain times the points of the predicted mask. This design is based on the insight that the model can already predict approximately correct masks and doesn't need much refinement. It ensures the model completes instance masks within a reasonable range.

The improvement in accuracy matches our intuition. The model is first trained to encode points and segment point clouds at a low level. Even though the pseudo labels we use in this step are over-segmented, the model can learn relatively good point feature representations and predict object parts. Then we use under-segmented masks to teach the model how to connect objects and predict complete instance masks.
\newcommand{\gray}[1]{\textcolor{gray}{#1}}
\newcommand{\green}[1]{\textcolor[RGB]{96,177,87}{#1}}
\newcommand{\fn}[1]{\footnotesize{#1}}\newcommand{\bk}[1]{\fn{(#1)}}
\newcommand{\gbf}[1]{\green{\bf{\fn{(#1)}}}}
\newcommand{\rbf}[1]{\red{\bf{\fn{(#1)}}}}
\newcommand{\bbf}[1]{\bf{\fn{(#1)}}}
\newcommand{\demph}[1]{\textcolor{Gray}{#1}}

\section{Experiments}
\paragraph{Implementation detablackils}
For point downsampling in preprocessing, we downsample the whole point cloud to the half of the number of original points and set $k_1=k_2=4$.
\paragraph{Datasets}
We evaluate our work on two publicly available indoor 3D instance segmentation datasets ScanNet~\cite{dai2017scannet} and S3DIS~\cite{armeni20163d}. The ScanNet dataset altogether contains 1613 scans, divided into
training, validation and testing sets of 1201, 312, 100 scans respectively. We use the 20-class benchmark provided by the  dataset. The S3DIS dataset contains 3D scans of 6 areas with 271 scenes in total. The dataset consists of 13 classes for instance segmentation evaluation. For unsupervised instance segmentation, we train on the training set. We report both evaluation results on the training set following UnScene3D~\cite{rozenberszki2023unscene3d} and the validation set.
\paragraph{Evaluation Metrics}
We use standard average precision as our evaluation metrics. AP$_{50}$ and AP$_{25}$ denote the scores with IoU thresholds of 0.5 and 0.25 respectively. AP denotes the average scores with IoU threshold from 0.5 to 0.95 with a step size of 0.05. We evaluate only instance mask AP values without considering any semantic labels.
\subsection{Main Results}
\begin{figure}
\begin{center}
\includegraphics[width=1\linewidth]{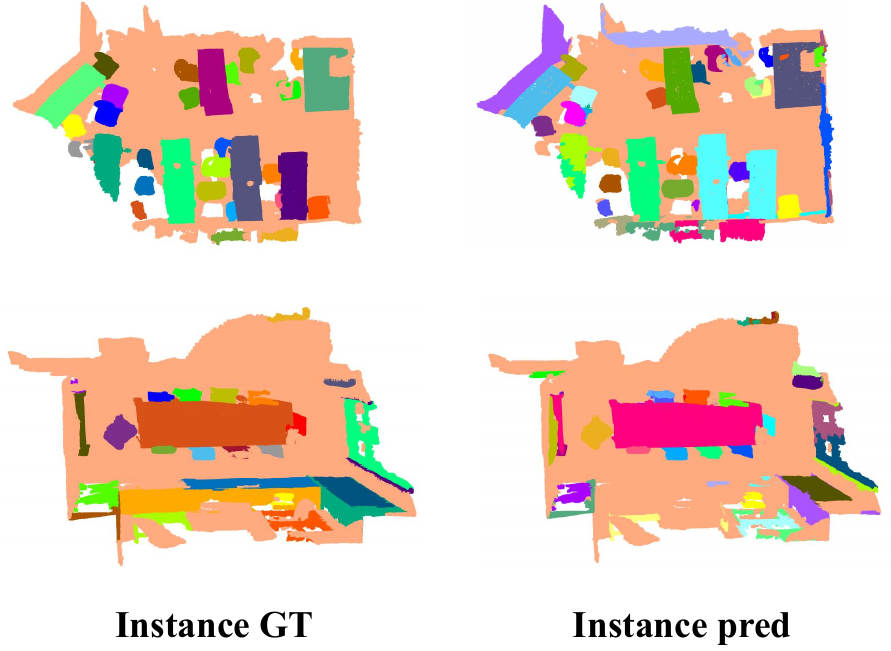}
\end{center}
   \caption{\textbf{Qualitative results on ScanNet.} FreePoint shows surprisingly good performance without any annotations.}
\label{fig:visualization}
\end{figure}

\paragraph{Unsupervised instance segmentation}
We mainly compare our work with a concurrent work~\cite{rozenberszki2023unscene3d}, which is a recently proposed unsupervised instance segmentation method for indoor 3D scenes. It operates on a basis of geometric oversegmentation to generate pseudo labels and refines them through multi-round self-training as many works. We also compare FreePoint with some traditional clustering methods including DBSCAN~\cite{ram2010density}, HDBSCAN~\cite{mcinnes2017accelerated} and a method originally proposed for outdoor autonomous vehicles~\cite{nunes2022unsupervised}. The visualization results are shown in Figure~\ref{fig:visualization}.

We report the result in Table~\ref{tab:class-agnostic}. It is worth noting that UnScene3D utilizes both 3D pretraining deep features and 2D pretraining deep features, while we only use the former. For a more fair comparison, we also show the result of UnScene3D which only uses 3D features from the same pretraining method CSC~\cite{Hou21CVPR} as FreePoint. Our method surpasses previous methods by a significant margin.

\begin{table}
\begin{center}
\begin{tabular}{lccccc}
\hline
\multirow{2}{*}{Method} &\multicolumn{2}{c}{Train set}&\multicolumn{2}{c}{Val set} \\ 
& AP & AP$_{50}$ & AP & AP$_{50}$\\
\hline
DBSCAN~\cite{ram2010density} &3.2&4.1&3.3&3.6\\
HDBSCAN~\cite{mcinnes2017accelerated}&1.6 &5.5 & 1.9 & 5.4 \\ 
Nunes et al.~\cite{nunes2022unsupervised}&2.3 &7.3 & 2.1 & 6.9\\
UnScene3D~\cite{rozenberszki2023unscene3d} &13.3 &- &- & -\\
UnScene3D*~\cite{rozenberszki2023unscene3d} &15.9&32.2 &- &- \\
\textbf{FreePoint (Ours)} & \textbf{21.4} & \textbf{38.7} & \textbf{18.9} & \textbf{36.4}\\
\hline
\end{tabular}
\end{center}
\caption{\textbf{Unsupervised class-agnostic instance segmentation} on ScanNet train split and validation split. We report average precision (AP) with different IoU thresholds. We mainly compare our method with some traditional clustering methods for point clouds and some recently proposed deep-learning-based methods. '*' means the method utilizes both 2D features and 3D features. '-' means the result is not provided by the original paper and we don't have access to the code to evaluate it by ourselves. Our method improves significantly over baselines.}
\label{tab:class-agnostic}
\end{table}
\paragraph{Fine-tuning on semantic instance segmentation}

\begin{table}
\begin{center}
\resizebox{\linewidth}{!}{
\begin{tabular}{lclll}
\hline
{}&Pre-train & AP & AP$_{50}$ & AP$_{25}$\\
\hline
\multirow{6}{*}{\rotatebox{90}{\emph{10\% masks}}}
& \demph{Train from scratch} & \demph{34.7} & \demph{47.6} & \demph{56.3} \\  
&Supervised & 36.9 & 50.1 & 55.5\\ 
&PointContrast~
~\cite{xie2020pointcontrast} & 36.1~\rbf{-0.8} & 49.4~\rbf{-0.7} & 56.8~\bbf{+1.3} \\
&DepthContrast~\cite{zhang2021self} & 36.8~\rbf{-0.1} & 49.0~\rbf{-1.1} & 57.3~\bbf{+1.8}\\
&CSC~\cite{Hou21CVPR} & 37.1~\bbf{+0.2} & 50.7~\bbf{+0.6} & 57.1~\bbf{+1.6} \\
&\textbf{FreePoint (Ours)} & \textbf{42.9}~\gbf{+6.0} & \textbf{54.6}~\gbf{+4.5} & \textbf{61.1}~\gbf{+5.6} \\
\hline
\multirow{6}{*}{\rotatebox{90}{\emph{20\% masks}}}
& \demph{Train from scratch} & \demph{44.1} & \demph{54.3} & \demph{61.1} \\  
&Supervised & 45.7 & 55.2 & 61.4\\ 
&PointContrast~
~\cite{xie2020pointcontrast} & 44.4~\rbf{-1.3} & 54.8~\rbf{-0.4} & 61.7~\bbf{+0.3} \\
&DepthContrast~\cite{zhang2021self} & 45.2~\rbf{-0.5} & 54.9~\rbf{-0.3} & 62.4~\bbf{+1.0}\\
&CSC~\cite{Hou21CVPR} & 46.3~\bbf{+0.6} & 56.4~\bbf{+1.2} & 61.5~\bbf{+0.1} \\
&\textbf{FreePoint (Ours)} & \textbf{47.4}~\gbf{+1.7} & \textbf{60.2}~\gbf{+5.0} & \textbf{65.9}~\gbf{+4.5} \\
\hline
\end{tabular}
}
\end{center}
\caption{\textbf{Supervised semantic instance segmentation} with limited instance masks. ``Supervised'' denotes the process of \textit{fully-supervised pre-training} on ScanNet, succeeded by fine-tuning on S3DIS. In contrast, other methods employ \textit{unsupervised pre-training}. The numerical values in brackets indicate the relative performance changes of unsupervised pre-training compared to their supervised counterparts.}
\label{tab:limited_masks}
\end{table}

\begin{table}
\begin{center}
\resizebox{\linewidth}{!}{
\begin{tabular}{lclll}
\hline
{} & Pre-train & AP & AP$_{50}$ & AP$_{25}$\\
\hline
\multirow{6}{*}{\rotatebox{90}{\emph{10\% scenes}}}
&\demph{Train from scratch} & \demph{30.1} & \demph{41.2} & \demph{52.2} \\  
&Supervised & 32.4 & 41.8 & 52.3\\
&PointContrast~\cite{xie2020pointcontrast} & 31.0~\rbf{-1.4} & 42.2~\bbf{+0.4} & 53.5~\bbf{+1.2} \\
&DepthContrast~\cite{zhang2021self} &32.2~\rbf{-0.2} & 41.5~\rbf{-0.3} & 53.7~\bbf{+1.4}\\
&CSC~\cite{Hou21CVPR} & 32.7~\bbf{+0.3} & 42.7~\bbf{+0.9} & 54.4~\bbf{+2.1} \\
&\textbf{FreePoint (Ours)} & \textbf{37.2}~\gbf{+4.8} & \textbf{48.1}~\gbf{+6.3} & \textbf{59.3}~\gbf{+7.0} \\
\hline
\multirow{6}{*}{\rotatebox{90}{\emph{20\% scenes}}}
& \demph{Train from scratch} & \demph{42.1} & \demph{49.5} & \demph{58.3} \\  
&Supervised & 44.8 & 51.7 & 59.6\\
&PointContrast~\cite{xie2020pointcontrast} & 43.7~\rbf{-1.1} & 50.8~\rbf{-0.9} & 60.5~\bbf{+0.9} \\
&DepthContrast~\cite{zhang2021self} & 44.0~\rbf{-0.8} & 51.6~\rbf{-0.1} & 62.1~\bbf{+2.5}\\
&CSC~\cite{Hou21CVPR} & 44.4~\rbf{-0.4} & 52.9~\bbf{+1.2} & 61.0~\bbf{+1.4} \\
&\textbf{FreePoint (Ours)} & \textbf{48.1}~\gbf{+3.3} & \textbf{56.6}~\gbf{+4.9} & \textbf{64.3}~\gbf{+4.7} \\
\hline
\end{tabular}
}
\end{center}
\caption{\textbf{Supervised semantic instance segmentation} with limited fully annotated point clouds. ``Supervised'' denotes the process of \textit{fully-supervised pre-training} on ScanNet, succeeded by fine-tuning on S3DIS. In contrast, other methods employ \textit{unsupervised pre-training}. The numerical values in brackets indicate the relative performance changes of unsupervised pre-training compared to their supervised counterparts.}
\label{tab:limited_scenes}
\end{table}

Since our work is unsupervised, it can also be seen as a pre-training pretext task. Apart from unsupervised class-agnostic instance segmentation, we further evaluate our work's performance as an unsupervised pre-training model. As shown in Table~\ref{tab:limited_masks}, FreePoint pre-training significantly outperforms other unsupervised pre-training methods~\cite{xie2020pointcontrast, Hou21CVPR,zhang2021self} by a large margin, and even suppress the supervised pre-training by 6.0$\%$ AP and 1.7$\%$ AP and when using 10$\%$ and 20$\%$ training masks respectively. 

We also compare the pre-training methods with different amounts of full-scene annotations. As shown in Table~\ref{tab:limited_scenes}, we conduct fine-tuning experiments with only limited scenes available. Our work can still achieve satisfactory results. FreePoint pre-training outperforms other unsupervised pre-training methods, and even the supervised pre-training by 4.8$\%$ AP and 3.3$\%$ AP, when using 10$\%$ and 20$\%$ full-scene annotations respectively.

\subsection{Ablation Study}
In this part, we conduct ablation experiments to show the effectiveness of each designed component.
\paragraph{Different feature representations}
 We explore results on different kinds of point feature representations. For features generated by various self-supervised pre-training encoders, we compare their performance in generating coarse masks and final instance segmentation results. Then we combine the best performer with traditional features and find the accuracy can be further improved. The comparison between different feature representations is shown in Table~\ref{tab:feature_representation}.

\begin{table}
\begin{center}
\begin{tabular}{lcc|cc}
\hline
Method & AP & AP$_{50}$ & AP & AP$_{50}$ \\
\hline
Traditional &6.3& 10.4& 10.3& 21.6\\
PointContrast~\cite{xie2020pointcontrast} & 7.6& 13.3& 15.7 &27.9 \\
CSC~\cite{Hou21CVPR} & 7.9 & 13.4& 16.5 & 30.8   \\
\textbf{FreePoint (Ours)}  & 8.5 & 15.3 &18.9  & 36.4  \\
\hline
\end{tabular}
\end{center}
\caption{\textbf{Different feature representation methods }for generating base masks. We report the accuracy of both base masks (left block) and final results (right block). Our strategy has the best performance.}
\label{tab:feature_representation}
\end{table}

\paragraph{Segmenting methods}
Owning relatively good feature representation, there are many existing ways to segment point clouds and generate coarse masks accordingly. We compare some methods including Supervoxel~\cite{papon2013voxel}, FreeMasks, a method proposed by ~\cite{wang2022freesolo}, and spectral~\cite{melas2022deep} methods. For each method, we adapt them to the ScanNet dataset and tune parameters to achieve good results as far as we can.

\begin{figure}
\begin{center}
\includegraphics[width=1.0\linewidth]{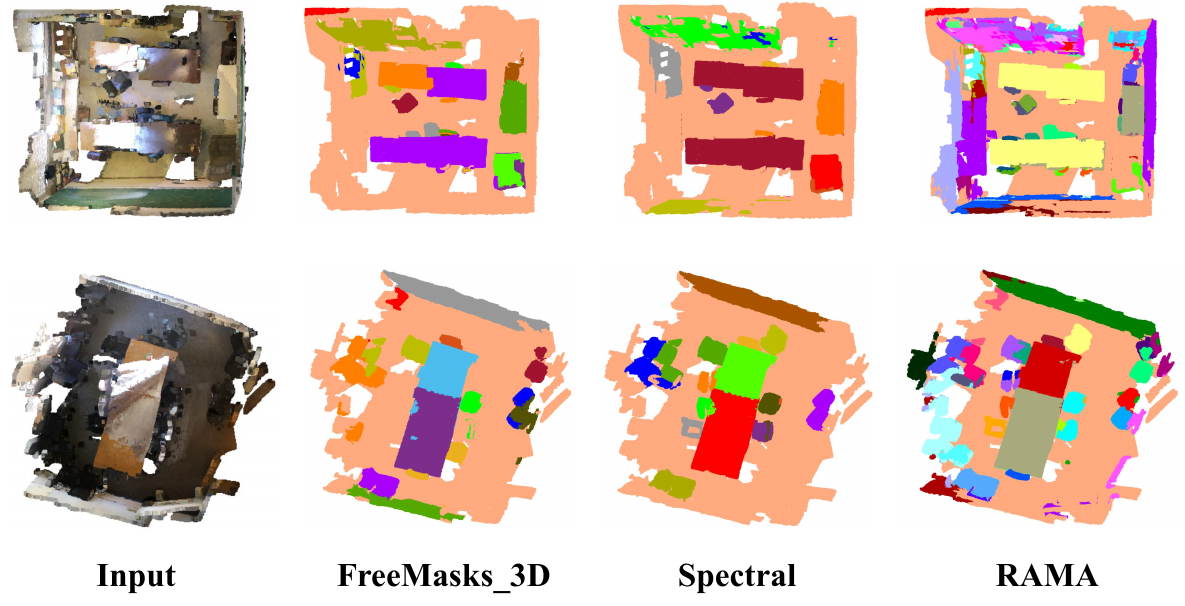}
\end{center}
   \caption{\textbf{Comparison with segmenting methods originally for 2D unsupervised instance segmentation.} Recent methods~\cite{wang2022freesolo, melas2022deep} for 2D unsupervised instance segmentation fail to deal with crowded and cluttered point cloud scenes due to their top-down mechanism. }
\label{fig:segmenting_methods}
\end{figure}

We observe that FreeMasks and spectral methods, which have proven successful in unsupervised object detection or segmentation tasks in the 2D field, fail to transfer to point clouds as shown in Figure~\ref{fig:segmenting_methods}. These two methods have two main defects due to their shared top-down mechanism. Firstly, they can only identify and localize partial objects in a crowded and cluttered 3D scene. Secondly, it is hard for these non-distance-based segmenting methods to distinguish different objects of the same semantic information even if they are far away from each other. The above two defects do not have much impact on some 2D images since they generally contain only one or a few dominant objects. But point cloud scenes are not this case. Point clouds usually have many similar objects in each scene, leading to unsatisfactory results. RAMA's bottom-up mechanism relieves the above problems in essence. We also explore the effectiveness of our id-as-feature strategy and the impact of the running times ${\bf{T}}$ of RAMA when adopting this strategy. For each method, we report the accuracy of coarse masks and final predictions. Results are shown in Table~\ref{tab:segmenting_methods}.    

\begin{table}
\begin{center}
\begin{tabular}{lcc|cc}
\hline
Method & AP & AP$_{50}$  & AP & AP$_{50}$ \\
\hline
Supervoxel~\cite{papon2013voxel} & 2.4 & 3.5 & 3.8 & 6.9 \\
FreeMasks\_3D~\cite{wang2022freesolo} & 2.9 & 3.2 & - & -\\
Spectral~\cite{melas2022deep} & 2.3 & 4.8 & - & - \\
RAMA  & 5.4 & 10.6 & 13.8 & 24.7 \\
RAMA-5  & 8.3& 14.7 & 17.6 & 35.0 \\
RAMA-10 & 8.5&  15.3 &18.9  &36.4  \\
\hline
\end{tabular}
\end{center}
\caption{\textbf{Segmenting methods.} We report the accuracy of both base masks (left block) and final results (right block). `-' means failing to converge. The number after RAMA is running times ${\bf{T}}$ for evaluation of our id-as-feature strategy.}
\label{tab:segmenting_methods}
\end{table}

\paragraph{Weakly-supervised learning design.}
To validate the effectiveness of our weakly-supervised design including different loss terms and undersegmentation-ignore method, we first evaluate the result of using fully-supervised loss(\textit{i.e.,} Dice loss and BCE loss) alone, discovering that directly adopting such loss leads to unsatisfactory results. We also find that only using our proposed weakly-supervised loss terms is even much worse than only using fully-supervised loss terms. This may be attributed to that terms for weak supervision contain too little information, unable to match low-quality predictions with ground truth at the early training stage. Each loss term and undersegmentation-ignore design are validated in Table~\ref{tab:loss}. 

\begin{table}
\begin{center}
\begin{tabular}{lccccc}
\hline
Method & AP & AP$_{50}$  \\
\hline
combination(default) & 18.9 & 36.4  \\
- w/o $\mathcal{L}_{mean}$ &  14.3 & 30.6 \\
- w/o $\mathcal{L}_{box}$ & 15.2 & 31.4 \\
- w/o $\mathcal{L}_{mean}$ and $\mathcal{L}_{box}$& 14.0 & 28.5   \\
- w/o $\mathcal{L}_{dice}$ and $\mathcal{L}_{BCE}$ & 7.8 & 15.7  \\
- w/o undersegmentation-ignore & 16.8 & 36.3  \\
\hline
\end{tabular}
\end{center}
\caption{\textbf{Weakly-supervised learning design.} Each design contributes to the final results.}
\label{tab:loss}
\end{table}

\paragraph{Training strategy 
}
As mentioned in section~\ref{segmenting}, we can generate coarse masks of different segmenting levels by changing parameters when running RAMA. Base masks are generally over-segmented while can identify and localize most objects in the scene. Therefore after training with the base masks, we further train the model with under-segmented masks with only a few epochs. In Table~\ref{tab:strategy} we report AP and AP$_{50}$ to evaluate our design's effectiveness.

\begin{table}
\begin{center}
\resizebox{\linewidth}{!}{
\begin{tabular}{lccccc}
\hline
Method & AP & AP$_{50}$ \\
\hline
base masks & 8.5 & 15.3 \\  
under-segmented masks &9.1  & 12.5 \\
train with base masks  &14.2  & 30.5  \\
train with under-segmented masks &6.4  & 13.8 \\
\textbf{Ours} & 18.9 &  36.4\\
\hline
\end{tabular}
}
\end{center}
\caption{\textbf{Training strategy.} Our two-step training strategy significantly improves the accuracy.}
\label{tab:strategy}
\end{table}

\section{Discussion and Conclusion}
In this work, we propose an effective framework FreePoint for unsupervised class-agnostic point cloud instance segmentation. FreePoint achieves satisfactory results compared with previous methods in this underexplored field, which proves this task is worthy of further exploration. In our experiment, we also find that top-down segmenting methods proposed in previous 2D unsupervised instance segmentation works fail to be directly adopted by point clouds as shown in Figure~\ref{fig:segmenting_methods}. Developing a novel unsupervised segmenting method for cluttered 3D indoor scenes may be promising. We hope our work can provide insights for future unsupervised point cloud learning works.
\paragraph{Acknowledgement} This work is supported by National Natural Science Foundation of China grants under contracts NO.62325111 and No.U22B2011. We would like to thank Ahmed Abbas for engaging in a discussion about the usage of the RAMA algorithm.

\clearpage
 {
     \small
     \bibliographystyle{ieeenat_fullname}
     \bibliography{main}
 }


\end{document}